\documentclass{article}

\usepackage{microtype}
\usepackage{graphicx}
\usepackage{subcaption}
\usepackage{booktabs} 
\usepackage{natbib}
\usepackage{xcolor}
\usepackage{commath}
\usepackage{xcolor}
\usepackage{tabularx,lipsum}
\usepackage{booktabs}
\usepackage{makecell}
\usepackage{hyperref}

\usepackage[accepted]{icml2020}

\icmltitlerunning{A Deep Learning Approach to Dst Index Prediction}
\newcolumntype{R}{>{\raggedright\arraybackslash}X}
\newcolumntype{W}{>{\centering\arraybackslash}X}
\begin{document}

\twocolumn[
\icmltitle{A Deep Learning Approach to Dst Index Prediction} 


\begin{icmlauthorlist}
\icmlauthor{Yasser Abduallah}{goo}
\icmlauthor{Jason T. L. Wang}{goo}
\icmlauthor{Prianka Bose}{goo}
\icmlauthor{Genwei Zhang}{goo}
\icmlauthor{Firas Gerges}{goo}
\icmlauthor{Haimin Wang}{to}

\end{icmlauthorlist}

\icmlaffiliation{to}{Institute for Space Weather Sciences, 
New Jersey Institute of Technology, University Heights, Newark, NJ 07102, USA}
\icmlaffiliation{goo}{Department of Computer Science, New Jersey Institute of Technology, Newark, NJ 07102, USA}

\icmlcorrespondingauthor{Jason T. L. Wang}{wangj@njit.edu}

\icmlkeywords{Machine Learning}

\vskip 0.3in
]



\printAffiliationsAndNotice{\icmlEqualContribution} 

\begin{abstract}
The disturbance storm time (Dst) index
 is an important and useful measurement in space weather research. 
 It has been used to characterize the size and intensity of a geomagnetic storm.
 A negative Dst value means that the Earth's magnetic field is weakened, which
happens during storms.
In this paper, we present a novel deep learning method,
called the Dst Transformer, 
to perform short-term, 1-6 hour ahead, forecasting of the Dst index based on the solar wind parameters
provided by the NASA Space Science Data Coordinated Archive.
The Dst Transformer combines 
a multi-head attention layer 
with Bayesian inference,
which is capable of quantifying both
aleatoric uncertainty and epistemic uncertainty when making Dst predictions.
Experimental results show that the proposed Dst Transformer outperforms
related machine learning methods
in terms of the root mean square error and R-squared.
Furthermore, the Dst Transformer can produce both data and model uncertainty 
quantification results, which can not be done by the existing methods.
To our knowledge, this is the first time that Bayesian deep learning 
has been used for Dst index forecasting. 
\end{abstract}

\section{Introduction}
Geomagnetic activities have significant impact on Earth.
They can disturb or damage telephone systems, power grid transmission systems and space satellites.
Geomagnetic activity modeling and forecasting has therefore been an important subject in space weather research. 
The main source of geomagnetic activity is solar activity.
The solar wind, which is a stream of charged particles released from the atmosphere of the Sun,
is considered as the medium through which the Sun exerts influence on Earth.
As a consequence, solar wind parameters such as the interplanetary magnetic field (IMF), total electric field,
solar wind speed and plasma temperature
are often used to model geomagnetic activities
and forecast geomagnetic indices.

The disturbance storm time (Dst) index is an important geomagnetic index.
It has been used to characterize the size and intensity of a geomagnetic storm. A negative Dst value means that the Earth's magnetic field is weakened, which 
happens during storms.
A storm is considered moderate when Dst is greater than $-50$ nT,
 intense when Dst is between $-50$ nT and $-250$ nT, or
 super when Dst is less than $-250$ nT~\citep{DST_StormCategories,MHCNNDstIndex2018SpWea..16.1882G}.
 
Many techniques have been developed to model and forecast the Dst index.
For example, \citet{Burton}
adopted differential equations to model the Dst index. 
The authors used solar wind parameters as the source of differential equations in their model.
\citet{Lundstedt-1994}
created the first Dst prediction model 
by employing a time-delay artificial neural network (ANN) with solar wind parameters as input. 
The authors performed 1-6 hour ahead predictions for the Dst index forecasting.
\citet{Bala} discussed another strategy by combining physical models and ANNs, 
along with parameters such as the solar wind velocity, IMF magnitude, and IMF clock angle.
\citet{Lazzus} employed a particle swarm optimization technique to train ANN connection weights 
to improve the accuracy of Dst index predictions. 

The above approaches mainly focused on single point predictions.
\citet{Chandorkar}
extended the above approaches by considering
probabilistic forecasting of the Dst index.
The authors used Gaussian processes (GP) to build autoregressive models to estimate Dst 1 hour ahead 
based on past Dst values, as well as the solar wind velocity and the IMF Bz component. 
Their technique generated a predictive distribution instead of single point predictions. 
However, the mean values of the forecasts are not as accurate as the forecasts produced by ANNs.
To improve GP's poor point prediction performance,
\citet{MHCNNDstIndex2018SpWea..16.1882G}
 built a Dst index prediction model by combining GP with a long short-term memory (LSTM) network.

In this paper, we present a novel Bayesian deep learning approach for performing 
short-term, 1-6 hour ahead, predictions of the Dst index. 
Our approach, called the Dst Transformer and denoted by DSTT,
combines a multi-head attention layer with Bayesian inference 
capable of handling both aleatoric uncertainty and epistemic uncertainty.
Aleatoric uncertainty, also known as
data uncertainty, measures the noise inherent in data.
Epistemic uncertainty, also known as model uncertainty, measures the
uncertainty in the parameters of a model \citep{UncertatintyComputervision10.5555/3295222.3295309}.
Thus, our work extends the aforementioned GP-based probabilistic forecasts, 
which can only handle model uncertainty, to quantify both data and model uncertainties
through Bayesian inference.
It is worth noting that Bayesian deep learning has also been used to mine solar images \cite{HaodiFibri2021}.
However, Dst values are time series data, not image data, and hence
the architecture of our Dst Transformer is totally different from the model architecture
developed by \citet{HaodiFibri2021}.

The contributions of our work are summarized below.
\begin{itemize}
\item Our DSTT model is the first to utilize the Transformer network to forecast the Dst index for a short-term period (i.e., 1-6 hours ahead).
\item This is the first study in which both data and model uncertainties are quantified when performing Dst index forecasting.
\item Our DSTT model outperforms closely comparable machine learning methods for short-term Dst index forecasting, 
as evidenced by performance metrics including the root mean square error (RMSE) and R-squared (R$^2$).
\end{itemize}

\section{Data}\label{sec:data}

\subsection{Data Source}
The Dst index measurements used in this study are provided by the NASA Space Science Data Coordinated Archive.\footnote{\url{https://nssdc.gsfc.nasa.gov}}
The data source provides other widely accessed data that are frequently used in solar wind analysis.
The data source is being periodically updated with Advanced Composition Explorer (ACE).\footnote{\url{https://omniweb.gsfc.nasa.gov}}
We used the Dst index data in the time period between January 1, 2010 and November 15, 2021. 
We selected the time resolution of the hourly average 
for the Dst index.
Following \citet{Ahmed2018PredictingDst},
we considered seven solar wind parameters, namely
the interplanetary magnetic field (IMF), 
magnetic field Bz component, 
plasma temperature, 
proton density, plasma speed, 
flow pressure, and electric field.
The total number of records in our dataset is 104,080.
The Dst index values in the dataset range from 77 nT to $-223$ nT.

\subsection{Data Labeling}
We divided our dataset into two parts: training set and test set. 
The training set contains 102,976
records from January 1, 2010 to September 30, 2021.
The test set contains 1104 records from October 1, 2021 to November 15, 2021. 
The training set and test set are disjoint.
The records are labeled as follows.
Let $t$ be a time point of interest and let $w$ be the time window ahead of $t$, 
where $w$ ranges from 1 to 6 hours for the short-term Dst forecasting studied here.
The label of the record at time point $t$
is defined as the Dst index value at time point $t + w$ for $w$-hour-ahead forecasting.
Each record in the training set has eight values including the seven solar wind parameter values and the label of the training record.
Each record in the test set contains only the seven solar wind parameter values; 
the label of each testing record in the test set will be predicted by our DSTT model.

\section{Proposed Method}\label{sec:methodology}
\subsection{Architecture of the DSTT Model}\label{sec:architecture}
\begin{figure}
 \centering
 \includegraphics[width=\columnwidth ]{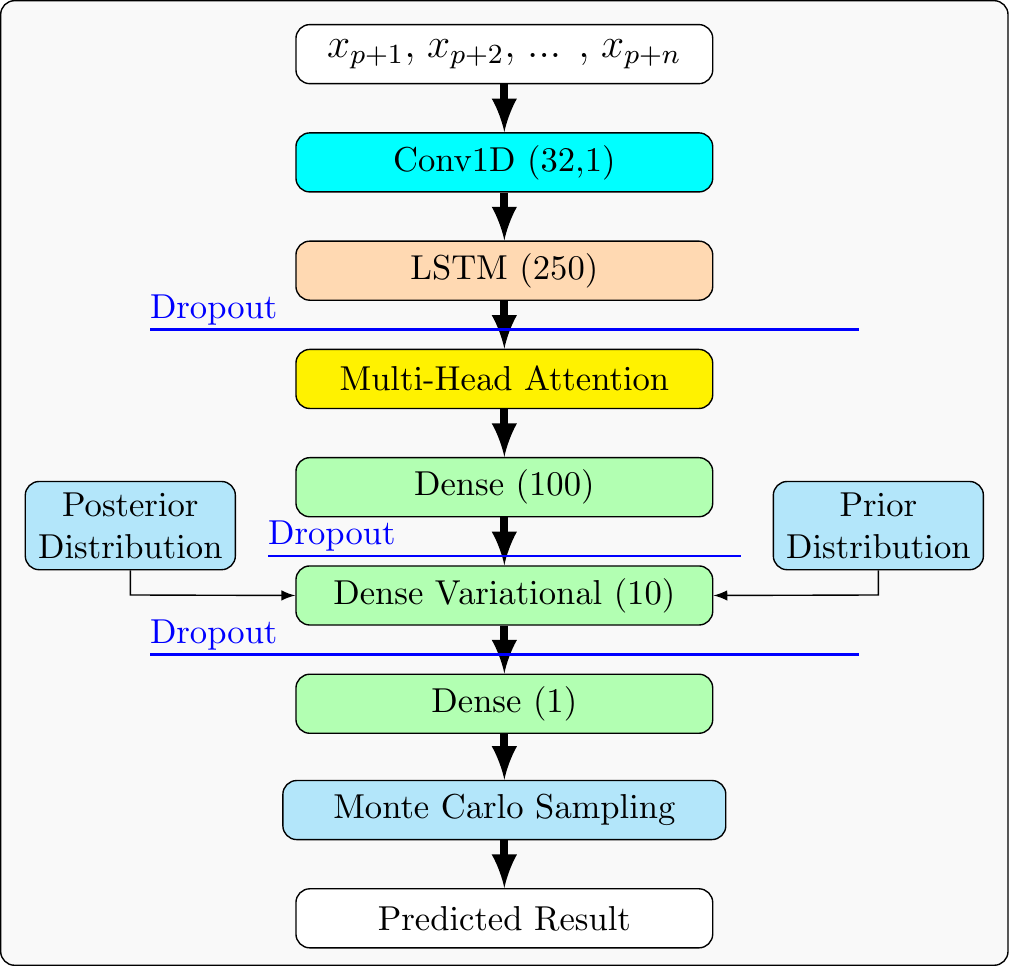}
 \caption{Architecture of our Dst Transformer (DSTT).}
 \label{fig:contextualarchitecture}
\end{figure}
Figure~\ref{fig:contextualarchitecture} presents the architecture of our DSTT model. 
DSTT is created using the tensorflow keras framework.\footnote{\url{https://www.tensorflow.org}}
We add multiple layers to DSTT to enhance its performance and improve its learning capability.
The model accepts as input non-overlapping sequences of records 
${x_{p+1}}$, ${x_{p+2}}$, $\ldots$ , ${x_{p+n}}$, 
where $n$ is set to 1024 in our study. 
Each sequence is passed to a one-dimensional convolution neural network (Conv1D) with 32 kernels
where the size of each kernel is 1.
Conv1D is well suited for sequential data;
it learns patterns from the input data sequence
and passes them to a long short-term memory (LSTM) layer 
that is configured with 250 LSTM units.
Combining Conv1D and LSTM layers has shown significant improvement in performance 
when dealing with sequential data such as time series \citep{TSINet_FLAIRS_Abduallah_Wang_Shen_Alobaid_Criscuoli_Wang_2021,faghihi_kalantarpour_2021}.
 LSTM hands the learned patterns down to a multi-head attention layer \citep{AttentionAllYouNeed10.5555/3295222.3295349}. 
The multi-head attention layer provides transformation on the sequential input of values to obtain distinct metrics of size $h$. 
Here, $h$ is the number of attention heads that is set to 3 and the size of each attention head is also set to 3
because a number greater than 3 caused overhead and less than 3 caused performance degradation.
The other parameters are left with their default values. 

Furthermore, we add custom attention to instruct the layers to focus and pay more attention to critical information of
the input data sequence and capture the correlation between the input and output by computing the weighted sum of the data sequence. 
 In addition, we add a dense variational layer (DVL) \citep{VaritionalLayer10.5555/3454287.3455600}
 with 10 neurons
 that uses variational inference \citep{Blei2016VariationalIA} to approximate the posterior distribution over the model weights. 
 DVL is similar to a regular dense layer, but requires two input functions that define the prior and posterior distributions over the model weights. 
 DVL allows our DSTT model to represent the weights by a distribution instead of estimated points.
 
 DSTT also includes multiple dense and dropout layers. 
 Each dense layer is strongly connected with its preceding layer where every neuron in the dense layer is connected with every neuron in the preceding layer. 
 Each dropout layer instructs the DSTT model to randomly drop a percentage of its hidden neurons 
 throughout the training phase to avoid over-fitting of training data. 

\subsection{Uncertainty Quantification}
 Quantifying uncertainty with a deep learning model has been used in many applications 
 such as medical image processing \citep{KWON2020106816},
 computer vision \citep{UncertatintyComputervision10.5555/3295222.3295309}, 
 space weather \citep{MHCNNDstIndex2018SpWea..16.1882G} 
 and solar physics \citep{HaodiFibri2021}.
Our proposed DSTT model contains a dense variational layer (DVL) that provides a weight distribution 
and multiple dropout layers that drop or turn off certain number of neurons during the training phase. 
Dropout is mainly used in deep learning to prevent over-fitting, 
where a trained model can be generalized for prediction instead of
fitting exactly against its training data.
With the dropout, the model's internal architecture is slightly different each time the neurons are dropped. 
This is an important behavior to the Monte Carlo (MC) class of algorithms 
that depends on random sampling and provides useful information \citep{MoteCarolDropout10.5555/3045390.3045502}. 
We use this technique 
to introduce a distribution interval of predicted values as demonstrated in Section ``Experiments and Results."

Specifically, to quantify the uncertainty with our DSTT model, we use a prior probability, $P(W)$, 
over the model's weights, $W$. During training, 
the seven solar wind parameter values and Dst index values, collectively referred to as $D$, are used to train the model. 
According to Bayes’ theorem,
	\begin{equation}\label{eq:bayestheorem}
		P(W|D) = \frac{P(D|W)\times P(W)}{P(D)}.
	\end{equation}
Computation of the exact posterior probability, $P(W|D)$, is
intractable \citep{HaodiFibri2021}, but we can use
variational inference \citep{Graves_VaritionalNIPS2011_7eb3c8be} to learn the variational distribution over the model’s weights
parameterized by $\theta$, 
$q_\theta(W)$, by minimizing the Kullback–Leibler (KL) divergence of
$q_\theta(W)$ and $P(W|D)$ \citep{Blei2016VariationalIA}. 
According to \citet{MoteCarolDropout10.5555/3045390.3045502}, a network with a dropout provides variational approximation. 
To minimize the KL divergence, we use the dense variational layer (DVL) shown in Figure \ref{fig:contextualarchitecture}
and assign the KL weight to $1/N$ where $N$ is the size of the training set \citep{MinKL6121942}. 
We use the mean squared error (MSE) loss function and the adaptive moment estimation
(Adam) optimizer \citep{Goodfellow_DeepLearningBookDBLP:books/daglib/0040158} with a learning rate of 0.0001
to train our model.
Let $\hat{\theta}$ denote the optimized variational parameter obtained by training the model;
we use $q_{\hat{\theta}}(W)$ to represent the optimized weight distribution.

During testing/prediction, our model utilizes the MC dropout sampling technique to produce probabilistic forecasting results,
quantifying both aleatoric and epistemic uncertainties.
Dropout is used to retrieve 
$K$ MC samples 
by processing the test data $K$ times \citep{MoteCarolDropout10.5555/3045390.3045502}. 
(In the study presented here, $K$ is set to 100.)
For each of the $K$ MC samples, a set of weights is randomly drawn from $q_{\hat{\theta}}(W)$.
For each predicted Dst value, we get a mean and a variance over the $K$ samples. 
Following \citep{KWON2020106816,HaodiFibri2021}, we decompose the variance into aleatoric and epistemic uncertainties. 
The aleatoric uncertainty captures the inherent randomness of the
predicted result, which comes from the input test data. 
On the other hand, the epistemic uncertainty comes from the variability of
$W$, which accounts for the uncertainty in the model parameters
(weights).

\begin{figure*}[ht!]
 \centering
 \includegraphics[width=2\columnwidth]{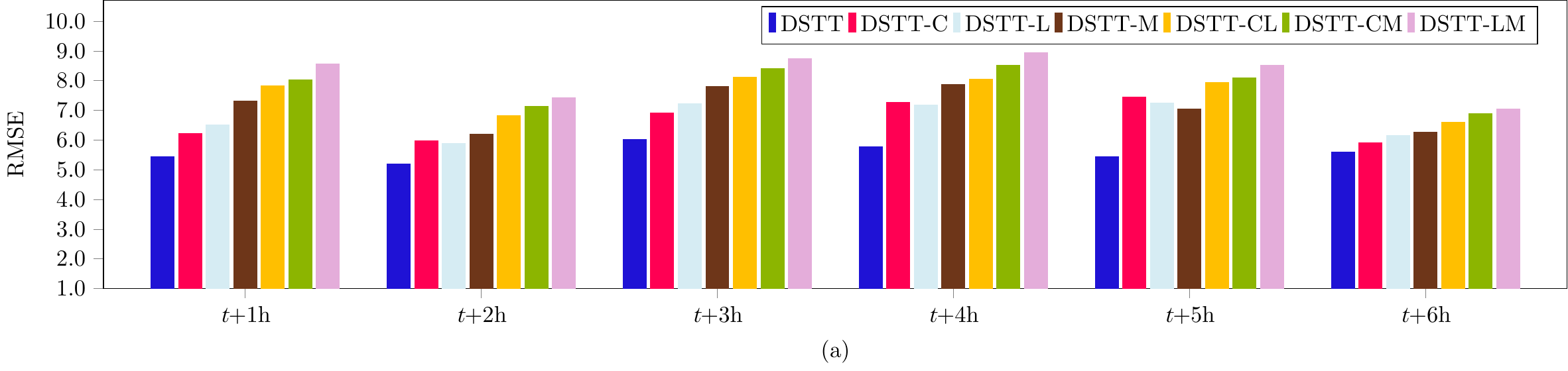}
 \includegraphics[width=2\columnwidth]{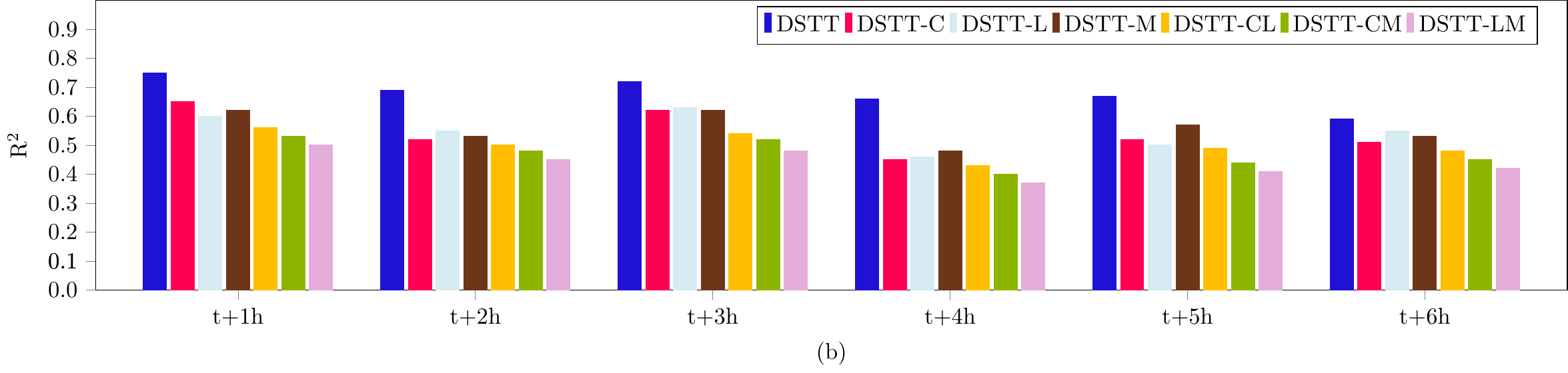} 
 \caption{Results of the ablation study.} 
 \label{fig:ablationtest}
\end{figure*}

\section{Experiments and Results}\label{sec:results}
\subsection{Performance Metrics}
We conducted a series of experiments
to evaluate our proposed DSTT model and compare it with closely related methods. 
The performance metrics used in our study are
the root mean square error (RMSE) \citep{TSINet_FLAIRS_Abduallah_Wang_Shen_Alobaid_Criscuoli_Wang_2021}
 and R-squared (R\textsuperscript{2}).
 
RMSE is calculated as follows:
	\begin{equation}\label{eq:rmse}
		\mbox{RMSE} = \sqrt{\frac{1}{m}\sum_{i=1}^{m}({y_i} - \hat{y_i})^2},
	\end{equation}
	where $m$ is the total number of testing records in the test set, 
	$\hat{y_i}$ ($y_i$, respectively) represents the predicted Dst index value (observed Dst index value, respectively) 
	at time point $i$. 
	The smaller the RMSE, the more accurate a method is.
	
R\textsuperscript{2} is calculated as follows:
	\begin{equation}\label{eq:r2squared}
		\mbox{R\textsuperscript2} = 1- \frac{\sum_{i}^m ({y_i} - \hat{y_i})^2}
		{\sum_{i}^m ({y_i} - \bar{y})^2},
	\end{equation}
 where $\bar{y}$ is the mean of the observed Dst index values.
 The larger the R\textsuperscript{2}, the more accurate a method is. 

\subsection{Ablation Study}
In this experiment, we performed ablation tests to analyze and evaluate the components of our DSTT model.
We considered six subnets derived from DSTT: DSTT-C, DSTT-L, DSTT-M, DSTT-CL, DSTT-CM and DSTT-LM. 
DSTT-C (DSTT-L, DSTT-M, DSTT-CL, DSTT-CM, DSTT-LM, respectively)
represents the subnet of DSTT in which we remove the Conv1D layer
(LSTM layer, multi-head attention layer, 
 Conv1D and LSTM layers,
 Conv1D and multi-head attention layers,
 LSTM and multi-head attention layers, respectively) 
 while keeping the remaining components of the DSTT network.
 For comparison purposes, we turned off the uncertainty quantification mechanism in the seven models.

Figure \ref{fig:ablationtest} presents the RMSE and R\textsuperscript{2} results of the seven models.
The $t+w$h, $1 \leq w \leq 6$, on the X-axis corresponds to
the $w$-hour ahead predictions of the Dst index based on the testing records in the test set.
 It can be seen from Figure \ref{fig:ablationtest} that our proposed full network model, DSTT, 
 achieves the best performance among the seven models. 
 DSTT-C captures the temporal correlation from the input data but it does not learn additional patterns and properties to strengthen the relationship between data records. 
 DSTT-L captures the properties from the data records but it lacks the temporal correlation information to deeply analyze the sequential information in the input data. 
 DSTT-M captures both the temporal correlation and additional properties, but it does not provide transformation on the sequential inputs
 to obtain distinct metrics to further strengthen the correlation between the predicted and observed Dst values. 
 Similarity, DSTT-CL, DSTT-CM, and DSTT-LM do not capture the combined patterns due to the removed layers. 
 As a consequence, the six subnets achieve worse performance than DSTT.
 It can be seen from Figure \ref{fig:ablationtest} that removing two layers yields worse results than removing one layer only.
 DSTT-LM yields the worst results, indicating the importance of including the LSTM and multi-head attention layers.
 The results based on RMSE and R\textsuperscript{2} are consistent.
 In subsequent experiments, we used DSTT due to its best performance among the seven models. 

\begin{figure*}[ht!]
 \centering
 \includegraphics[width=2\columnwidth]{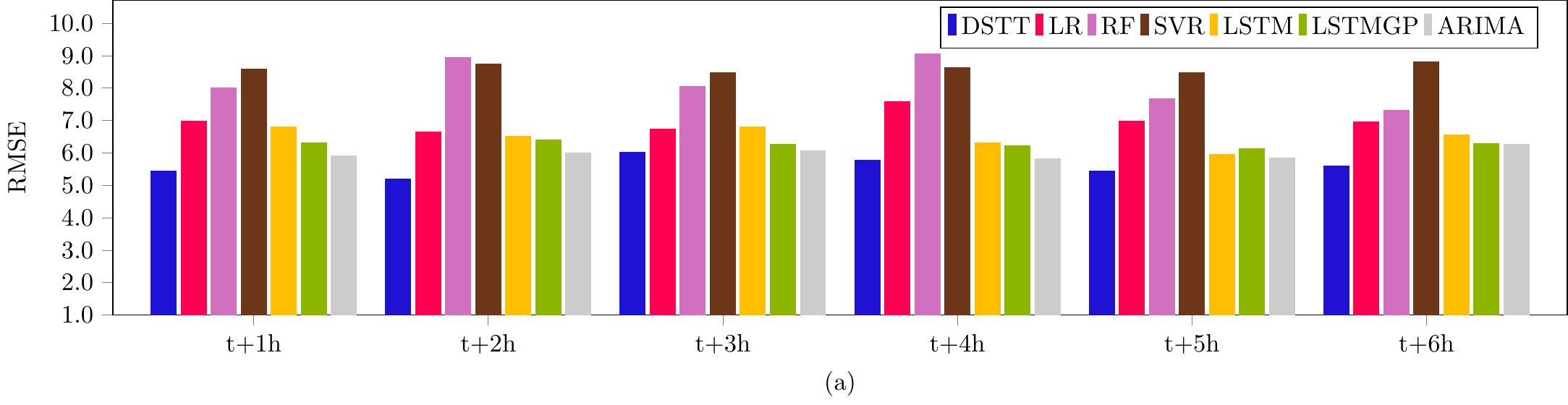}
 \includegraphics[width=2\columnwidth]{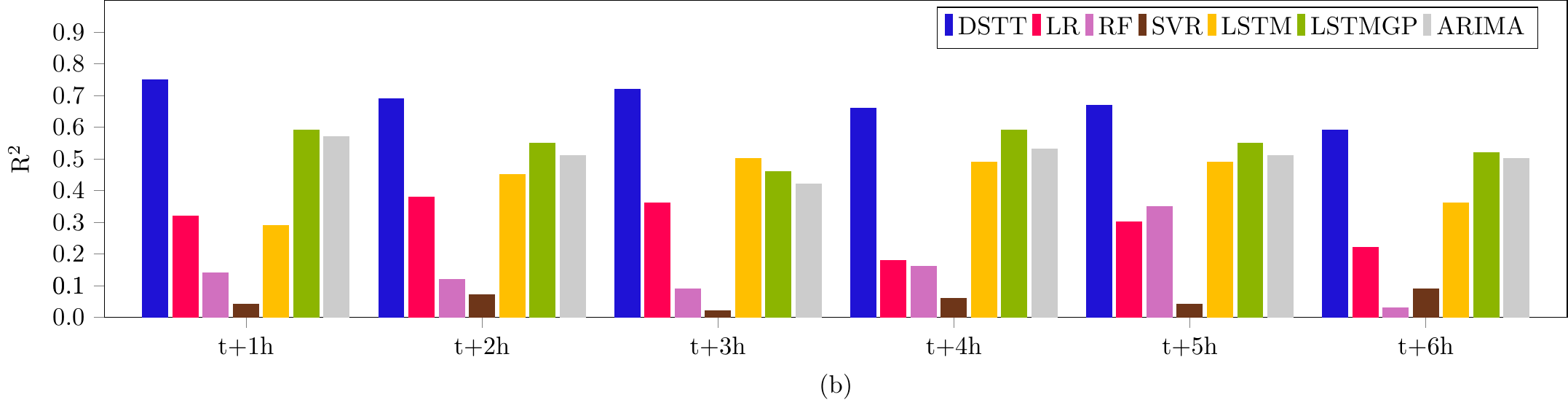}
 \caption{Performance comparison of seven Dst forecasting methods.}
 \label{fig:performancemetrics}
\end{figure*}

\subsection{Comparison with Related Methods}
In this experiment, we compared the Dst Transformer (DSTT) with six closely related machine learning methods including 
 linear regression (LR),
random forests (RF),
 support vector regression (SVR),
auto regressive integrated moving average (ARIMA)~\citep{ARIMA2022E3SWC.33600034T},
 long short-term memory (LSTM),
and the method developed by \citet{MHCNNDstIndex2018SpWea..16.1882G}, 
 which combines LSTM with Gaussian processes (GP) and is
 denoted by LSTMGP.
 Because the six related methods do not have the ability to quantify both data and model uncertainties, 
 we turned off the uncertainty quantification mechanism in our DSTT model when performing this experiment.

Figure \ref{fig:performancemetrics} 
presents the RMSE and R\textsuperscript{2} results of the seven methods:
DSTT, LR, RF, SVR, ARIMA, LSTM and LSTMGP.
It can be seen from the 
 figure that DSTT achieves the best performance, producing the most accurate predictions, among the seven methods
in terms of both RMSE and R\textsuperscript{2}.
The deep learning methods including DSTT, LSTM and LSTMGP as well as ARIMA mostly perform better than
the traditional machine learning algorithms including RF, LR and SVR.

\subsection{Uncertainty Quantification Results}
\begin{figure*}
 \centering
 \includegraphics[width=1\columnwidth]{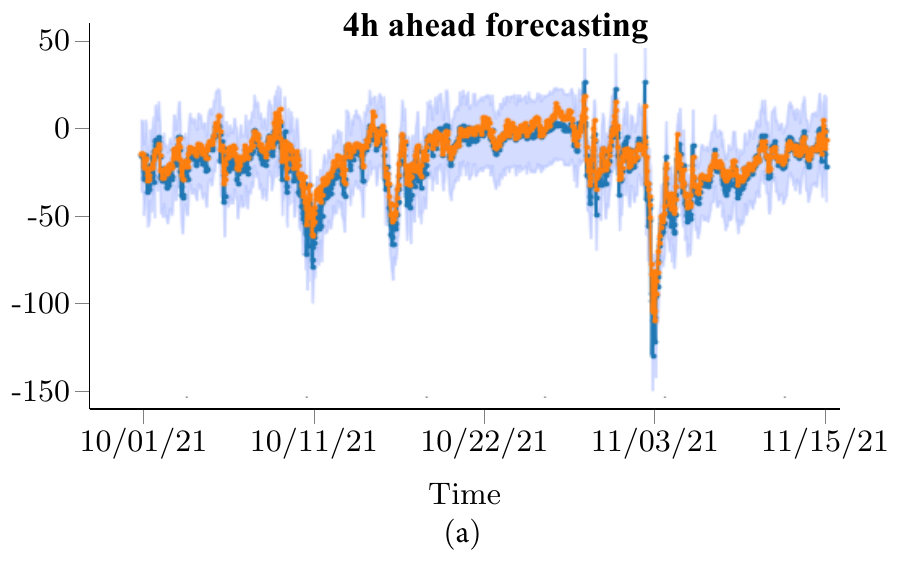}
 \includegraphics[width=1\columnwidth]{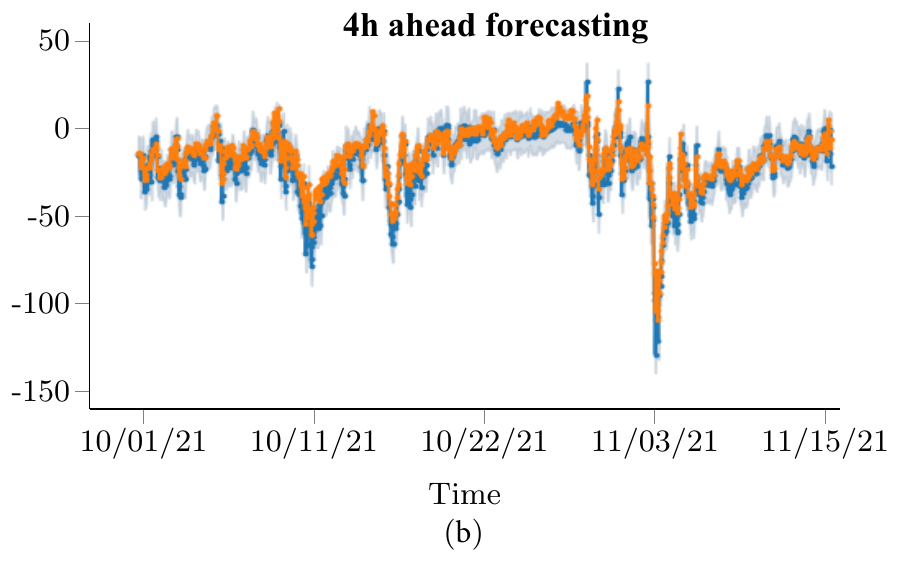} 
 \caption{Uncertainty quantification results produced by our Dst Transformer (DSTT).}
 \label{fig:uncertaintyquantification}
\end{figure*}

Figure~\ref{fig:uncertaintyquantification} shows 
uncertainty quantification results produced by our DSTT model 
on the test set.
Due to space limitation, we only present the results obtained from
4-hour ahead predictions of the Dst index.
In Figure~\ref{fig:uncertaintyquantification},
the orange lines represent observed values of the Dst index (ground truth)
while the blue lines represent the predicted values of the Dst index.
The light blue region in Figure~\ref{fig:uncertaintyquantification}(a) represents aleatoric uncertainty (data uncertainty).
The light gray region in Figure~\ref{fig:uncertaintyquantification}(b) represents epistemic uncertainty (model uncertainty).

It can be seen from Figure~\ref{fig:uncertaintyquantification} that
the blue lines are reasonably close to the orange lines, indicating the good performance of our DSTT model, 
which is consistent with the results in Figure \ref{fig:performancemetrics}.
 Figure~\ref{fig:uncertaintyquantification} also shows that 
the light gray region is much smaller than the light blue region,
indicating that the model uncertainties are much smaller than the data uncertainties.
Thus, the uncertainty in the predicted result is mainly due to the noise in the input test data.
Similar results were obtained from other $w$-hour, $1 \leq w \leq 6$, $w \not= 4$, ahead predictions of the Dst index.
We note that when $w$ is longer than 4, the prediction performance starts to degrade. This is understandable given that we are trying to predict a Dst index value that is farther away from the input 
test record.

\section{Discussion and Conclusions} \label{sec:conclusions}
The disturbance storm time (Dst) index is an important and useful measurement in space weather research,
which is used to understand the severity of a geomagnetic storm. 
The Dst index is also known as the measure of the decrease in the Earth’s magnetic field. 
In this paper, we present a novel deep learning model,
called the Dst Transformer or DSTT,
to perform short-term, 1-6 hour ahead predictions of the Dst index.
Our empirical study demonstrated the good performance of the Dst Transformer and
its superiority over related methods.

Our experiments were based on the data collected in the period between January 1, 2010 and November 15, 2021. 
The training set contained hourly records from January 1, 2010 to September 30, 2021.
The test set contained hourly records from October 1, 2021 to November 15, 2021. 
To avoid bias in our findings, we performed additional experiments using 10-fold cross validation (CV).
For the CV tests, we used the original data set described above 
and another data set ranging from
November 28, 1963 to March 1, 2022 
that has 510,696 records. 
In addition, we generated synthetic data with up to 1.2 million records to further assess the performance and stability of our DSTT model. 
With the 10-fold CV tests,
the data was divided into 10 approximately equal partitions or folds.
The sequential order of the data in each fold was maintained.
In each run, one fold was used for testing and the other nine folds together were used for training.
There were 10 folds and hence 10 runs.
We computed the performance metrics including RMSE and R\textsuperscript{2} for each method studied in the paper in each run.
The means and standard deviations of the metric values over the 10 runs were calculated and recorded.
Results from the 10-fold CV tests were consistent with those reported in the paper.
Thus we conclude that the proposed Dst Transformer (DSTT) is a feasible machine learning method 
for short-term, 1-6 hour ahead predictions of the Dst index.
Furthermore, our DST Transformer can quantify both data and model uncertainties 
in making the predictions, which can not be done by the related methods.

Our work focuses on short-term predictions of the Dst index by utilizing solar wind parameters.
These solar wind parameters are collected by instruments near Earth and are suited for
short-term predictions of the geomagnetic storms near Earth \citep{MHCNNDstIndex2018SpWea..16.1882G,Ahmed2018PredictingDst}.
When using the solar wind parameters to perform long-term (e.g., 3-day ahead) predictions of the Dst index,
the accuracy is low.
In future work, we plan to perform long-term predictions of the Dst index
by utilizing solar data collected by instruments near the Sun.
The solar data reflects solar activity, which is the source of geomagnetic activity.
We plan to extend the Bayesian deep learning method described here to mine the solar data for performing long-term Dst index forecasts.

\section{Acknowledgments}
We acknowledge the use of NASA/GSFC's Space Physics Data Facility's OMNIWeb service and OMNI data.

\bibliographystyle{icml2020}

\end{document}